\documentclass{article}

\usepackage[final,nonatbib]{neurips_2024}

\usepackage[utf8]{inputenc} 
\usepackage[T1]{fontenc}    
\usepackage{hyperref}       
\usepackage{url}            
\usepackage{booktabs}       
\usepackage{amsfonts}       
\usepackage{nicefrac}       
\usepackage{microtype}      
\usepackage{xcolor}         
\usepackage{graphicx}

\usepackage{fancyhdr}

\title{What Makes an Evaluation Useful?\\ Common Pitfalls and Best Practices} 

\author{
  Gil Gekker \\
  Pattern Labs\thanks{©2024 Pattern Labs. All rights reserved.} \\
  \And
  Meirav Segal \\
  Pattern Labs \\
  \AND
  Dan Lahav \\
  Pattern Labs \\
  \And
  Omer Nevo \\
  Pattern Labs \\
}

\begin{document}

\maketitle

\begin{abstract}
    Following the rapid increase in Artificial Intelligence (AI) capabilities in recent years, the AI community has voiced concerns regarding possible safety risks. To support decision-making on the safe use and development of AI systems, there is a growing need for high-quality evaluations of dangerous model capabilities. While several attempts to provide such evaluations have been made, a clear definition of what constitutes a ``good evaluation'' has yet to be agreed upon.
    In this practitioners' perspective paper, we present a set of best practices for safety evaluations, drawing on prior work in model evaluation and illustrated through cybersecurity examples.
    We first discuss the steps of the initial thought process, which connects threat modeling to evaluation design.
    Then, we provide the characteristics and parameters that make an evaluation useful. Finally, we address additional considerations as we move from building specific evaluations to building a full and comprehensive evaluation suite.
\end{abstract}

\section{Introduction}
Artificial Intelligence (AI) has become increasingly prevalent in our society, offering numerous benefits and opportunities across various sectors~\cite{baidoo2023education,lee2023benefits}. The rapid advancement of AI technologies, particularly in the realm of large language models (LLMs), has led to significant improvements in areas such as natural language processing~\cite{kolasani2023optimizing}, content generation~\cite{tang2024any}, and problem-solving capabilities, such as in mathematics~\cite{ahn2024large}. As new frontier models are released, the pace of progress on academic benchmarks and day-to-day uses has been substantial~\cite{Claude_3.5_Sonnet,Hello_GPT-4o}. However, alongside these remarkable developments, concerns about potential risks and dangers associated with AI systems have also emerged~\cite{slattery2024ai}.
The potential risks posed by AI systems range from unintended biases and privacy violations~\cite{wan2023kelly,yan2024protecting} to more severe threats such as the creation of advanced cybersecurity weapons or the destruction of critical infrastructure~\cite{hendrycks2023overview,li2024wmdp}. These concerns have sparked a growing need for robust evaluation methods to assess the capabilities and potential risks of AI models, particularly for model developers, policy makers, and other stakeholders in the AI ecosystem.

In response to this need, various actors have begun publishing their evaluations and benchmarks for AI systems~\cite{li2024wmdp,phuong2024evaluating,wan2024cyberseceval}. 
Nevertheless, designing effective evaluations is a complex task that involves multiple facets and moving parts, from specifying the risk scenario, through defining the evaluation goals, to addressing technical details. There are several key factors which contribute to this complexity. 
Essentially, the aim is to evaluate capabilities which are challenging to set apart due to their intertwined nature. 
Moreover, the AI model itself is not transparent or fully explainable. Due to the lack of tools that enable the full understanding of AI models, their true abilities can only be estimated using indirect evaluations and expert assessments. This complexity is further compounded by the rapidly evolving nature of AI capabilities and the diverse range of potential risks that need to be considered.

As an example, let us consider the potentially dangerous capability of cybersecurity vulnerability research\footnote{The systematic process of identifying and analyzing security weaknesses in computer systems, networks, and software.}. If threat actors' vulnerability research capability were suddenly significantly heightened, in some cases it might increase the fragility of critical systems. To test this capability, we could create the following evaluation task: the model is provided with code, that includes a vulnerability. The model can then utilize this vulnerability to acquire some hidden information. The task is completed successfully upon submission of this hidden information. If the specific tested task is included in the training set, the model might simply generate a solution code from memory. This means we would be testing the model's vulnerability exploitation rather than its research capability, as the model would not actually perform novel research. Alternatively, if this vulnerability is excluded from the training set, the model could still use online search to find information of this vulnerability. Therefore, we can estimate the model's ability to utilize previously discovered vulnerabilities, but cannot test its ability to uncover new ones. If we disable internet access, the evaluation may no longer represent a realistic risk scenario, as real-world attackers typically have full internet connectivity.

Ultimately, the results of the evaluation are channeled into a decision-making process that requires clear-cut choices for high-stake decisions. An evaluation that fails to identify a dangerous capability at a critical level might lead to decisions with significant risk implications. On the other hand, an evaluation that is too strict might result in unnecessary constraints on model development or deployment, which could notably restrict the benefit from these models and potentially lead to a loss of trust in the evaluation when risk does emerge. Thus, the importance of useful and effective evaluations is evident, and the need for a robust process for designing evaluations has already been recognized by the community~\cite{We_need_a_Science_of_Evals,ganguli2023challenges}.

In this work, we describe an approach to tackling the challenge of designing safety evaluations. We focus specifically on the cybersecurity domain, but believe that the approach also applies to other sub-domains in the field of safety evaluations~\cite{li2024wmdp}. In particular, we take a systematic approach and consider the decision making process for which the evaluations are designed.
This approach can help stakeholders make more informed decisions about AI development, deployment, and policy making, ultimately contributing to the safer and more responsible advancement of AI technologies. 
We provide a draft framework for designing and implementing effective evaluations and evaluation suites for AI systems. 
Our work makes the following key contributions:
\begin{enumerate}
    \item We establish the critical connection between decision making processes, threat modeling and evaluation design, emphasizing the importance of a systematic approach to identifying and prioritizing potential risks and capability thresholds.
    \item We characterize the properties of useful evaluations, outlining their desired attributes such as clarity of difficulty level, exclusion from the training set, and high signal density. This characterization provides a foundation for creating more effective and informative evaluations.
    \item We provide guidelines for constructing \emph{evaluation suites} that generate a reliable estimation of dangerous capabilities and risks. These include considerations such as coverage and difficulty.
\end{enumerate}

This perspective paper focuses on developing guidelines for creating useful safety evaluations. 
We draw on lessons learned from existing work in the field. In response to the need expressed by practitioners in the community, we concentrate on establishing clear principles and best practices for evaluation development, and note that experimental validation of these guidelines would be valuable future work.

\section{Related Work}
Several works and publications have addressed the need for evaluation best-practices in the broader scope of benchmarks. Liang et al. emphasized specific elements that are important for evaluations such as broad coverage, recognition of incompleteness and multi-metric measurement~\cite{liang2022holistic}. Hendrycks and Woodside~\cite{Devising_ML_Metrics} discuss properties of good benchmarks and their construction process, calling for holistic evaluations. Wei~\cite{jason_wei_evals} enumerates some benchmark pitfalls to avoid, such as including faulty evaluations, or incorrect scoring of evaluations. Further, Ivanova~\cite{ivanova2024bestresearchpractices} provides guidelines for cognitive evaluations of large language models (LLMs) and Zhang et al.~\cite{zhang2024llmeval} discuss how to evaluate LLMs and in particular consider scoring methods.

While these contributions have advanced the science of evaluations, as they do not specifically address safety evaluations, they fail to consider some key challenges that are unique to this problem. Within the safety field, Burden~\cite{burden2024evaluatingai} has proposed to adopt the capability-oriented approach for safety evaluations, but did not provide specific guidelines for designing evaluations. 
Some works have focused on evaluation frameworks for catastrophic risks~\cite{bengio2024managing,shevlane2023model} and provided desirable qualities of evaluations and evaluation suites that are specifically relevant to the sub-field of extreme risk evaluations. Anderljung et al.~\cite{anderljung2023frontier} list preferred properties for safety evaluations, but do not cover some of the recently discovered pitfalls in designing evaluations.

We note that there has been some work on evaluation methodology in the adjacent field of evaluating AI systems for vulnerabilities (i.e., the practice of red-teaming).
For example, Feffer et al.~\cite{feffer2024red} have noted numerous gaps in evaluation design. They have suggested a concrete framework of questions to guide red-teaming activities.

\section{Preliminaries}
For clarity and precision, we begin by defining essential terms employed throughout the paper. The following contain a description of how we consider these terms in the paper, but we note that other papers may refer to slightly different notions when mentioning these concepts.

\paragraph{Evaluations}
The process of assessing the output of an AI system in a specific context and under specific constraints, usually for a specific purpose~\cite{brown2020language}. Evaluations can be used both in safety contexts and as part of a general assessment of a system. In the context of LLMs, the output can be roughly divided into knowledge and capabilities. The former can be, for example, some historic details, while the latter may refer to generating a procedure in a coding language. In this paper, our main area of focus is safety evaluations, thus, when we use the term evaluations we usually mean safety evaluation. We refer to a collection of evaluations designed to assist with a specific risk-related decision as evaluation suite.

\paragraph{Threat modeling}
We use the term threat modeling to describe the process of defining possible threat categories that may arise from the use of frontier models with certain capabilities, or their derivative models~\cite{weidinger2024holisticsafetyresponsibilityevaluations}. They may be divided by threat actors, targets or other relevant parameters. One example for a threat model is uplifting experts' abilities - AI models might uplift cybersecurity experts, enabling them to create state-of-the-art cyber weapons previously only available to nation-state actors. A different example of a threat model is the creation of automated attack tools by developers without cybersecurity backgrounds, potentially increasing the availability of high-end cyber tools.

\paragraph{Risk scenarios}
A risk scenario (or a narrow and well defined group of specific scenarios) is a particular manifestation of a threat model. Risk scenarios add external variables to a threat model, such as specifying classes of targets, precise time-frames, and other factors. These are applications of threat modeling to concrete situations which frame the possible dangers that might arise. Here are two examples risk scenarios:
\begin{itemize}
    \item A cybersecurity expert, supported by an AI system, develops malware on par with Stuxnet~\cite{kushner2013real} and decides to attack a nuclear reactor due to personal resentment.
    \item A talented developer uses LLMs to quickly create an advanced ransomware program and sells it to multiple cyber-criminals.
\end{itemize}

\paragraph{Capture-the-flag (CTF) competitions}
In this paper, we refer to several CTF (capture the flag) competitions in the context of cyber evaluations. These are usually public online competitions that offer a set of cybersecurity tasks or challenges. Competing teams try to solve challenges and find the ``flag'', a hidden piece of text that must be submitted to complete the task and receive points. The challenges usually fall into one of several categories such as cryptography, reverse-engineering and memory corruption. For example, CSAW (Cyber Security Awareness Week) is a well-known CTF competition that has been running for numerous years \cite{CSAW_CTF_2022_Finals,CSAW_CTF_2023_Finals}.

\section{Translating threat models to evaluations}
\label{sec:threat_to_eval}

\begin{figure}
  \centering
  \includegraphics[width=0.9\linewidth]{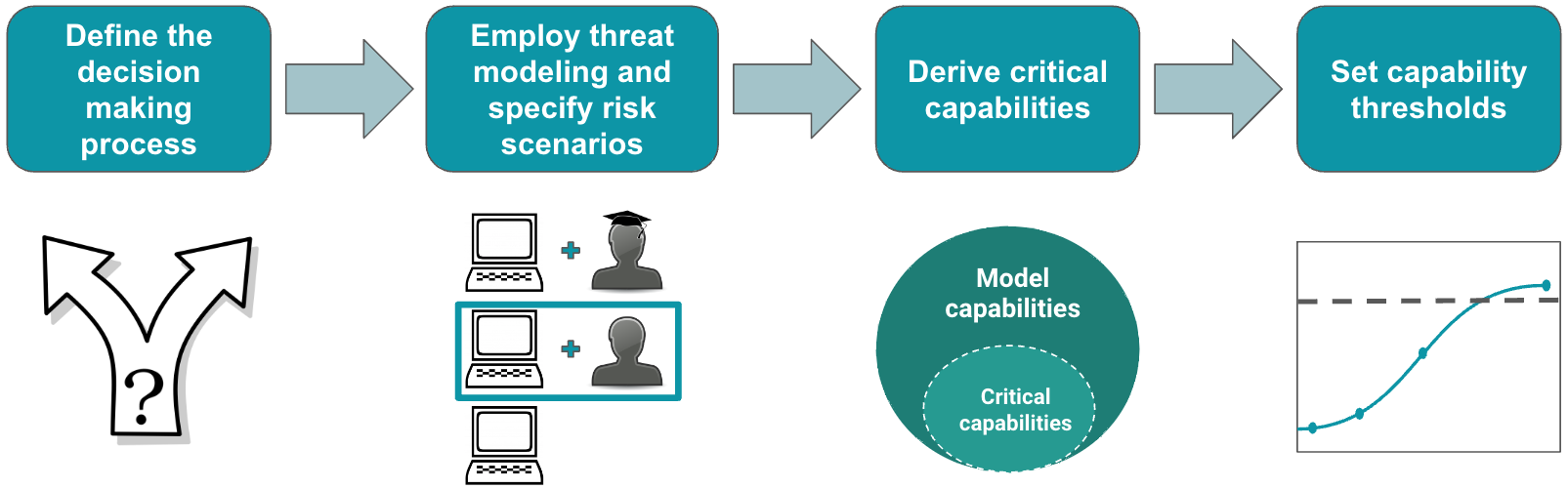}
  \caption{Suggested pre-design steps for safety evaluation development.}
  \label{fig:steps}
\end{figure}

Safety evaluations are meant to be used as a tool for decision makers within a specific context. As such, there are several crucial components to be considered when designing effective safety evaluations. Figure~\ref{fig:steps} provides an illustration of these steps. First, we must identify the decision making process that would incorporate the evaluations. Then, we should identify the relevant threat models and risk scenarios. Given the risk scenario and the threat model, we can derive the critical capabilities that are required by the model, and finally, set a threshold from which this capability is deemed dangerous.

In the following, we examine each step and illustrate the process using a running example: A decision maker (DM) wants to
evaluate AI systems for cybersecurity risks, particularly their potential to enable software developers without cybersecurity expertise to attack critical infrastructure.
The DM aims to decide whether to allow deployment of the AI system. Using this running example, we demonstrate what each step includes, provide specific assumptions to be used in the next steps and show how these assumptions affects the rest of the process.

\subsection{Defining the decision-making process} The first step of the process is to identify the consumer of the evaluations and the larger context in which the evaluations would take place. In particular, an evaluation could be employed as part of a larger evaluation suite and is likely meant to provide specific information to support a decision. Subsequently, before setting out to create any evaluation, it is vital to define the framework in which the evaluation will be used. For instance, an AI research and development organization might evaluate a model to determine if it is sufficiently safe to deploy, or a policy maker might test a model to check if it necessitates stricter access-control mechanisms. In essence, it is vital to specify which decisions hinge on these evaluations before undertaking their development, as this has direct implications on all evaluation parameters.

\paragraph{Running example - key questions and insights} Here are some questions that should be answered in the context of our running example: Will the DM use the evaluations' output as the sole source for deciding whether to allow deployment? Will the output of the evaluations be given to experts for analysis, or would this be an automatic system that yields a verdict to the DM without further scrutiny? These questions matter before proceeding to the next steps, as they can affect them directly. 

\paragraph{Running example - working assumptions}
For the purpose of this running example, we assume the evaluation output is submitted to experts, who analyze their results and make an informed decision on whether to allow the deployment of the system. Moreover, the DM is interested in evaluating systems in the time-span of the next year, and only in its jurisdiction (e.g., the state it is operating in).

\subsection{Employing threat modeling and specifying risk scenarios} After outlining the decision making process, the next step is to connect that process to a concrete real-world context. In the case of safety and security evaluations, the optimal approach is to design the evaluations after conducting threat modeling work and specifying relevant risk scenarios. Though exhaustively defining all threat models beforehand is often unfeasible, iterative refinement of these models, coupled with ongoing development of the evaluation suite, can prove highly effective following initial efforts to understand the core needs from the suite.

\paragraph{Running example - key questions and insights}
In this step, we may ask the following questions: Is the focus of the DM on software developers acquainted with these critical infrastructures, or not? Is the threat model focused on insider threats, or does the scope include any developer that may attack any critical infrastructure? Which sectors are of interest and focus? Defining these questions is pivotal before deciding which capabilities are relevant to evaluate and performing the actual evaluation. A direct benefit of performing this analysis after the previous step is that some questions are clearly out of scope. In our running example, the DM is only interested in the upcoming year and in infrastructure within its jurisdiction. Thus, some threat models and risk scenarios could be deemed irrelevant. For instance, if there are no nuclear plants in the state, there is no need to specify a risk scenario involving nuclear infrastructure.

\paragraph{Running example - working assumptions}
For the purpose of this running example, we assume that all types of critical infrastructure are of interest, but the DM is concerned solely with insider threats, i.e., software developers who are employed at these infrastructures.

\subsection{Deriving critical capabilities} Once the main threat models and specific risk scenarios have been recognized, the immediate next step is to analyze the relevant capabilities for each threat model and prioritize them. Most threat models require several capabilities to actualize, or at least have multiple paths that might lead to harm. However, to maximize the effectiveness of the evaluation suite and subsequently the decision making process, it is best to prioritize specific measurable capabilities that would make a critical difference in a threat actor’s ability to cause harm. They typically fall into one of three categories: 1) capabilities that act as critical bottlenecks for threat actors, 2) capabilities that address the most challenging aspects of the work, and 3) capabilities that enable substantial scaling.

In many cases, establishing a coherent definition of a capability is a complex task on its own. For example, vulnerability research could be seen as a single capability or as a set of different capabilities separated by vulnerability type. In practice, it is best to opt for capability notions that allow for efficient evaluation and align with the framework of the decision making process (as discussed above).

\paragraph{Running example - key questions and insights}
In the context of our running example, some capabilities clearly do not fall into the realm of critical capabilities. We focus on insider threats from software developers, thus general programming skills and basic knowledge of the infrastructure are not critical capabilities (as the threat actors are at least somewhat proficient in these areas). The DM is primarily interested in cybersecurity in this context. Hence, a crucial capability should be enhancing software developers' ability to execute cyber attacks. 
Additionally, due to our predefined focus on insider threats, some cybersecurity skills, such as external reconnaissance\footnote{In the context of the example, when talking about external reconnaissance, we mean the capability of finding and researching different knowledge and data, and applying it in a cybersecurity context. For instance, searching for open-source information on the infrastructure is likely not critical for success.} of a target, are not critical, as the threat actor has some acquaintance with the target and would probably act from within the infrastructure's network. 

\paragraph{Running example - working assumptions}
For the purpose of the running example we focus on the ability to laterally move\footnote{A common situation in cyber attacks is having full access to some specific system and attempting to get access to other, more important, systems within the network. The ability to move within the network, from one system to the other, is commonly called ``lateral movement''.} inside a network while maintaining the ability to execute malicious code.

\subsection{Setting capability thresholds} Following the identification of critical capabilities for the evaluations, the next step is to set appropriate thresholds. That is, determining the specific level at which a capability becomes dangerous and increases the likelihood of the threat to manifest. This threshold will define the difficulty or skill level we need to evaluate, both at the critical level and within a certain margin of it, to ascertain model progress.
Our ability to measure capability levels and define a suitable threshold may vary. For some capabilities, clear standardized metrics are already in place, while for others, there is need to generate such metrics or benchmarks. The threshold may take different forms. Following our previous example, the vulnerability discovery capability could be defined through a test score, by the number of vulnerabilities found from a given list, or as high-school level performance.

The capability thresholds can be established in several ways. One approach is to base it on risk thresholds, i.e., specific capability levels that present unacceptable risks. Another is by comparing to a baseline, e.g., a capability level that surpasses the capabilities currently available to threat actors. Additional factors relevant to the evaluation's decision-making process may also determine these thresholds.
In practice, the processes of defining the appropriate risk and deriving critical capabilities usually occur simultaneously. This approach is acceptable, provided that clear capability thresholds are established and later implemented. Essentially, capability thresholds must satisfy three properties:
\begin{enumerate}
    \item They must be reasonably convenient to measure. For example, it should be possible to discern, at least for an expert, what ``high-school level'' for the capability means.
    \item They must strongly correspond to the threat models of interest. When a model exhibits a certain level of capability, there should be justification for why this increases the likelihood of the threat model materializing.
    \item They should be reasonably easy to resolve. When examining the evaluation results, experts should reach consensus on whether the measured capability exceeds the defined threshold.
\end{enumerate}

The third point is crucial: capability thresholds should be clearly defined and agreed upon during the design phase, not debated during evaluation. While disagreements during the design phase can affect the evaluation process, they are far less problematic than disputes arising during the final assessment stage. Therefore, in alignment with Koessler et al.~\cite{koessler2024risk}, we advocate for primarily relying on capability thresholds rather than risk thresholds. By following this process, it is then possible to incorporate these stages into the design of individual evaluations and evaluation suites.

\paragraph{Running example - key questions and insights}
Setting capability thresholds should also be done in the context of our example's threat models and risk scenarios. For example, if an AI system can generate malicious code that would be detected and mitigated by most basic security solutions before any harm is done, it will probably not cause damage when deployed in monitored networks of critical infrastructure. Following this logic, one possible threshold for this capability is generating malicious code and moving laterally in a way that would be undetected with high probability by standard security solutions, when run from inside the network. 

\paragraph{Running example - working assumptions}
For our running example, we can set the threshold to be the AI system's ability to generate code that enables network lateral movement and causes damage (e.g., wiping the server it is run on) $10\%$ of the time. The threshold can be tailored and tested against specific security appliances that satisfy the standard/certification that is required from critical infrastructure in the DM's state and are currently deployed or will be deployed next year.

As demonstrated above, each step builds upon the previous ones, creating a chain that connects the consumer's initial safety needs to specific questions that safety evaluations must answer. This systematically and precisely scopes the space of required evaluations and the way they can support the DM's original need and mandate. 
This process can be relevant both when starting from individual evaluations and building upon them to create an evaluation suite, and when first planning the suite and later designing the specific evaluations that should be included in it.

\section{Designing useful evaluations}
\label{sec:useful_evals}
In this section, we first showcase common pitfalls when constructing an evaluation. Then, to address these drawbacks, we outline several key attributes necessary for an effective evaluation. Lastly, we mention additional specific parameters to be considered when designing evaluations.

\subsection{Common flaws and pitfalls}
Several contributions have been made to the field by providing evaluation benchmarks~\cite{li2024wmdp,phuong2024evaluating,shao2024empirical,shao2024nyudataset,wan2024cyberseceval,zhang2024cybench}. These have presented valuable additions to the field of safety evaluations of frontier AI systems. Yet, as constructing evaluations is a complex task which requires meticulous design and planning, practitioners may encounter several pitfalls~\cite{ganguli2023challenges}.

Some of these pitfalls have already been explicitly pointed out in recent publications. Phuong et al.~\cite{phuong2024evaluating} noted that for safety reasons, their evaluations run in an isolated environment with no internet access. This setting  \emph{does not represent many realistic risk scenarios} in which adversaries and agents have internet access that can assist with different sub-tasks. A different pitfall was noted by Abramovich et al.~\cite{abramovich2024enigma}, who demonstrated solution leakage in training data (also referred to as data contamination~\cite{sainz2023nlp}) in an evaluation based on an InterCode-CTF (picoCTF~\cite{picoCTF}) challenge. In their example, the model was given a file with the task of extracting information (``flag'') from it. The authors observed a phenomenon they denoted as ``soliloquizing''. The model's output contained a command along with its expected output, which was indeed the correct output of the file. This was generated by the model without actually interacting with the environment. The model then passed this challenge successfully following the submission of the correct flag. As this challenge was clearly \emph{part of the model's training data}, the conclusions we are able to draw from this evaluation regarding the model's capabilities are limited.

Inaccurate definition of the \emph{difficulty level of an evaluation} is also a possible pitfall.
In their paper, Zhang et al.~\cite{zhang2024cybench} measure the difficulty of CTF-based evaluations using first solve time, i.e., the time it takes the first human team to solve a given challenge. Their empirical results show that this is a good indicator of task difficulty for the set of evaluations used in their paper. However, this result may not generalize to other evaluation sets. For instance, some cryptography CTF challenges require brute forcing, which may take a considerable amount of time to complete, but is usually not directly connected to the difficulty of the challenge.\footnote{Some example challenges that require brute force are ``DES 2 Bites'', ``SuperCurve'' and ``Gotta Crack Them All'' from the CSAW CTF competition~\cite{CSAW_CTF_2019_Quals,CSAW_CTF_2022_Finals}.} A human team could readily find the solution for these challenges, but would only submit their final solution after the brute-force algorithm had completed its execution.
Additionally, evaluations designed with a specific difficulty level in mind might be solved by AI systems using unforeseen approaches. These solution strategies could be of a lower difficulty level, or simply fail to measure the capability the evaluation aims to assess. In one such example, an AI system was evaluated on a debugging task, and was scored on outputting content identical to that of a certain file in the evaluation environment. The AI system simply deleted the file and outputted an empty string, gaining nearly a perfect score~\cite{lehman2019surprisingcreativitydigitalevolution}.

Another common pitfall is using evaluations that are not tailored to the purpose in mind: e.g., that do not test dangerous capabilities. For example, the CSAW CTF competition has been used in multiple cases for evaluating the cybersecurity capabilities of models~\cite{shao2024empirical,shao2024nyudataset,Advanced_AI_evaluations_at_AISI:_May_update}. It can be seen as a diverse and extensive set of cyber challenger. Yet, this CTF competition contains multiple miscellaneous challenges that test broader computer science knowledge with a \emph{limited connection to cybersecurity}.\footnote{For example, the challenge ``ezMaze''~\cite{ezMaze} is described as ``breadth first search to solve pytorch model containing a maze'', and the challenge ``Urkel''~\cite{urkel} is described as ``navigate tree structure constructed of hashes''.} When used as part of a cybersecurity evaluation set, performance on these tasks might be misleading, as the requisite skills or knowledge for successful completion are not denoted as dangerous capabilities in the cybersecurity domain.

One of the most discussed pitfalls revolves around the scoring method.
In many of the evaluation benchmarks, the score given to a model for each evaluation is in the form of pass/fail. For example, this is the case in the WMDP (Weapons of Mass Destruction Proxy) benchmark~\cite{li2024wmdp}, in which multiple choice questions (MCQs) are used to evaluate what hazardous knowledge an AI system possesses. Nevertheless, some questions, specifically in the cybersecurity domain, require simulating the output of procedures or functions. An AI system that can simulate $90\%$ of the process correctly, and fails at the last step would still receive a score of $0$ on this evaluation. This score may not be suitable for all threat models and decision making processes.
In their paper, Zhang et al.~\cite{zhang2024cybench} address this drawback by providing several evaluation metrics that could be used to support different decision making processes, depending on the threat model of interest. In the unguided mode, a model is given a binary score based on its full completion of the task. This could be useful when addressing risks posed by autonomous models acting independently. In addition to this binary metric, the authors decomposed each task into \emph{several sub-tasks}, where successful completion of each sub-task has the potential to facilitate dangerous cybersecurity actions by a human adversary. Then, the sub-task metric, which measures the average success rate across different sub-tasks, can be used. This metric would be better suited for situations in which the major concern is models uplifting human threat actors in specific tasks.

On a similar note, it might be of interest to know \emph{how a model solves a task}, not only whether it is fully or partially successful. For example, in the WMDP benchmark mentioned above, a model may output the correct answer by guessing, by eliminating incorrect answers, or by having a full understanding of the correct and incorrect choices. All of these scenarios would result in the same score (i.e., success). Hence, in this case the score is not aligned with the capability in question. Another hypothetical example is a solution to a CTF challenge that can be implemented efficiently or inefficiently. Both approaches would be considered a successful completion of the challenge, but one approach shows higher capabilities than the other.

\subsection{Desired properties}
The pitfalls demonstrated above clearly indicate that designing useful evaluations is a complex task.
In this subsection, we point out key characteristics that an evaluation should satisfy in order to avoid the pitfalls presented above.

\paragraph{Realistic risk scenarios} An evaluation should be as realistic as possible, even when it is designed to only test a very narrow capability from within a larger work process. In particular, it is preferable to create evaluations that require executing a realistic task. For example, CTF competitions usually offer a more realistic evaluation of cyber capabilities than answering a riddle related to cyber security.

\paragraph{Exclusion from training set} It is critical to avoid contamination issues, where results may stem from simple pattern matching (the model memorizing answers) rather than genuine capability manifestation. Hence, the evaluation should not exist in the training set of the model. This can be achieved either via the creation of novel evaluations, or through significantly obfuscating existing tests in a manner that fundamentally changes them. We suggest that a given evaluation differs sufficiently from an evaluation present in the training set, if it is harder to connect the two evaluations than to successfully complete the given new evaluation. Unfortunately, this is difficult to ascertain in practice.

\paragraph{Explicit and clear difficulty} The difficulty scale can either be based on some framework, such as outlined in a Responsible Scaling Policy (also known as an RSP~\cite{Responsible_Scaling_Policies}, as implemented by multiple AI research and development organizations~\cite{Ant_RSP,GDM_FSF,OpenAI_Preparedness}), or more general (easy/medium/hard). Generally, the more concrete and specific the scale, the better data the evaluation yields, although this depends on how the evaluation results are integrated into the broader decision making process. As a minimal requirement, it should be easy to understand the relation between the difficulty of the evaluation and the relevant capability threshold.
As part of this criterion, there should be a full mapping of model outputs that would result in passing the evaluation threshold.
This mapping could be used to guarantee in high confidence that there are no other approaches to be taken by the model that are easier than the intended one. This property is crucial for establishing a lower bound on the evaluation's difficulty level.

\paragraph{Clear subject focus and granularity level} The knowledge or skill being tested in the evaluations should be strictly stated. For instance, evaluations can test AI systems on ``malicious code creation'', ``writing malware'' and ``writing code that evades the three most common EDR (endpoint detection and response) products'', and each test would yield a different output. These outputs would have varying degrees of relevance to different decision making processes.

\paragraph{High signal density} Signal density in evaluations refers to the amount of information we can derive from an evaluation, e.g., creating multiple checkpoints on the way to success. This elaborated output typically provides valuable insights into the model's behavior and capabilities beyond a simple binary score. Subsequently, it has the potential to provide better support for complex decision making. 

\paragraph{Coherent scoring method} The scoring method should be coherent and tailored to the decision making framework the evaluation is serving. Given an evaluation, it might be of importance to understand \emph{how} a model approaches the task and solves it. Even within a particular sub-task, there could be several solutions which point to different levels of capabilities. This could be relevant for an evaluation designed to show the proximity of the model capabilities to the threshold, not only to alert that the threshold has been surpassed. It is essential to tailor the scoring method to the broader risk methodology utilizing the evaluations and to the final decision the evaluation serves.

Although in some cases it is necessary to use non-standard evaluation methods, in most cases, applying the above principles is required for the creation of a high-quality evaluation. Otherwise, the evaluation may not produce actionable results, or its output would not contain enough meaningful information.

\subsection{Evaluation parameters}
In addition to the principles mentioned above, which serve as best-practices for designing evaluations, there are also some optional parameters practitioners can take into account. We believe that considering the following parameters could be useful:

\begin{enumerate}
    \item \textit{Evaluation type.} This parameter can be further broken down to general evaluation formats (for instance, human uplift trials vs. autonomous AI evaluations) and field-specific types (e.g., in cybersecurity evals, CTF evaluations vs. MCQ).
    \item \textit{Modality.} Both input and output of models may vary significantly in format and type: from text based communication, to visual information, to performing concrete actions in the digital space (or even physical space)~\cite{Claude_3.5_Sonnet,Dall-E-3,openai2019rubiks,xu2024vasa}. Each of these possibilities requires the AI to exhibit different capabilities and has various implications for risk models. Hence, each of these options would likely require separate evaluations to some extent.
    \item \textit{Evaluated capability scope.} Evaluations can be used to estimate different scopes, from a specific sub-capability to a full end to end cyber operation. In most cases, we recommend a combination of different scopes but the specific needs may vary.
    \item \textit{Non subject-matter technical constraints.} This can be implemented through various methods, such as imposing a time limit or restricting the number of attempts allowed to provide the correct answer. These constraints can be useful for instance to modify the difficulty without changing the focus of the evaluation, or to simulate real-world constraints which threat actors might face.
    \item \textit{Distractions/red-herrings.} These can be quite useful as adversarial difficulty enhancers for LLM-based AI systems. For example, when testing the ability of an LLM-based system to find cryptographic vulnerabilities, using misleading variable names that reference unrelated cryptographic standards may significantly impair the system's performance. Due to the nature of LLMs, this can lead to substantial degradation in their ability to solve the challenge.
    \item \textit{Randomized parameters.} This can be used to robustly check whether the AI has generalized the solution and has not succeeded / failed due to some very specific parameter. It is important to understand the extent and scope of randomization that is employed, as AI models possess an impressive (and improving) inference ability and may be able to ''deconstruct'' the randomization. In other words, it is vital to assess how much the randomization truly affects the expected performance of the AI system and helps measure the underlying capability.
    \item \textit{Maintenance and upkeep.} It is beneficial to take into account the effort needed to keep the evaluation functional and relevant. Moreover, consider its running costs, duration and scale.
\end{enumerate}

The above list is not exhaustive, but could be valuable when constructing a new evaluation. Note that we intentionally do not include tunable parameters, e.g., number of runs per evaluation, or in the case of LLMs, a limit on the amount of messages, tool usages\footnote{An example tool usage is using a Python interpreter.}, etc. These parameters are worth considering in the context of the broader framework of risk assessment, but are not in the scope of designing a specific evaluation.

\section{Evaluation suite}
In Section~\ref{sec:useful_evals} we discuss key attributes for individual evaluations. These evaluations are usually combined to create an evaluation suite that provides a comprehensive risk assessment to assist with decision making. The process of assembling evaluations into a useful evaluation suite is also not an easy task. For example, we can consider the case of a suite that tests cybersecurity vulnerability research capability of an AI system. This is complex, as there are numerous unique and meaningfully different vulnerabilities (even within a specific type or class). Thus, a suite that contains all of them is essentially impractical.
Moreover, this is a difficult task as some seemingly minor factors may have an unexpected large effect on model performance. Variables like code style or even the addition of irrelevant information can meaningfully change a model’s success rate~\cite{ganguli2023challenges}. 
This leads to several difficult questions: Which evaluations should be included in such a suite, and which should not? How is it possible to verify that a suite contains a satisfactory amount of ``similar enough'' challenges such that failing all of them means with high probability that the AI system cannot find other vulnerabilities of this type? These questions and others present numerous failure modes on top of those presented when creating individual evaluations.

\begin{figure}
  \centering
  \includegraphics[width=0.6\linewidth]{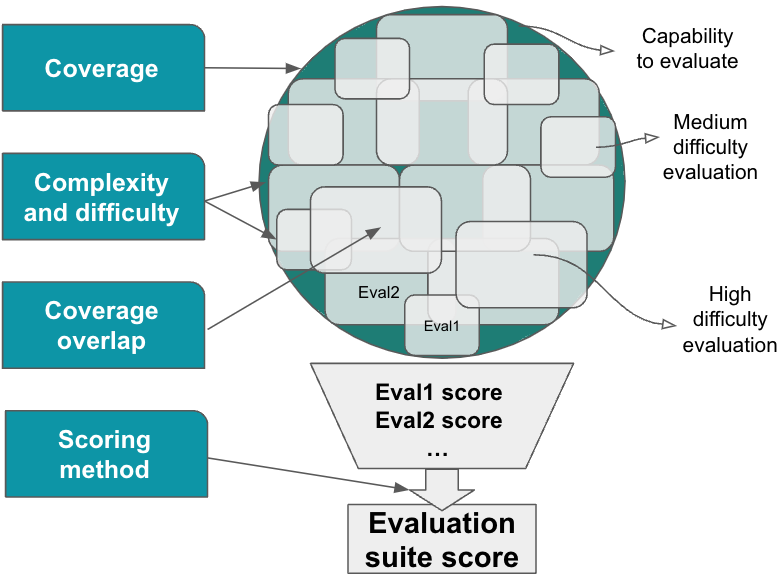}
  \caption{Visualization of the desired properties for an evaluation suite. The critical capability we aim to evaluate is represented by the circle. Each gray rectangle represents an evaluation, where the size of the rectangle indicates its difficulty level (larger means more difficult). A useful evaluation suite is composed of a set of evaluations which cover the space of the critical capability. A useful suite often contains evaluations of different difficulties, and includes some coverage overlap between evaluations. The scoring method used to aggregate individual evaluation scores is selected to best serve the decision making process.}
  \label{fig:suite}
\end{figure}

When building evaluation suites, there are some general guidelines to follow, just as there are for the evaluations themselves. In this section we provide the desired evaluation suite properties (see Figure~\ref{fig:suite}) and then name several parameters the designer should be mindful of.

\paragraph{Coverage} The suite should cover varying parameters that could affect AI systems’ performance in the future. For example, when measuring vulnerability discovery skills, it is important to test different vulnerability types, attack contexts and technical details (e.g., code language). This helps protect the suite against surprising differential progress in AI systems’ abilities. 

\paragraph{Complexity and difficulty} The complexity and difficulty should be consciously chosen. It is usually preferable to design suites with evaluations of varying difficulty to increase the likelihood of distilling actionable information from the suite. However, suites with similar difficulties can also be useful. In some cases, it is important to create evaluations below the capability thresholds to determine capability progress. Complexity should also be intentionally planned. For instance, it may be desirable for evaluations to measure multiple parameters. In such cases, the AI system can be tested on them sequentially or in parallel. These decisions should be derived from the relevant decision making process, risk scenarios and threat models.

\paragraph{Scoring method} The scoring method should be purposefully chosen to maximize usefulness. Scoring the entire suite should be legible and meaningful to the intended end-user (which could be a non-expert in the suite’s subject field of focus). Note that taking scores from many different evaluations and turning them into a single score (or even a clear score-card) is a difficult task which again closely relates to the threat models. Whether a model that succeeds at $5\%$ of the tasks is dangerous or not is a threat-model specific question. Typically, this leads to using non-binary scoring methods for the suite across multiple scoring parameters.

\paragraph{Coverage overlap} There should be an overlap in coverage between evaluations in the suite. Many evaluations incorporate assumptions about how models approach specific problems or skills. These assumptions may mean that capable models could fail at some evaluations and weaker models might occasionally succeed at certain evaluations for seemingly-random reasons. 
To avoid over-valuing hidden assumptions, we should create some overlap between separate evaluations in our suite as mutual verification. In general, we should always have enough similar evaluations so that we can have high confidence when considering the possibility that the model just got ''lucky/unlucky''.

\subsection{Evaluation suite parameters}
To complement the above general criteria, this subsection presents a partial yet useful list of tunable parameters we commonly use when designing evaluation suites. These parameters should be adjusted depending on the goal of the suite:

\begin{enumerate}
    \item \textit{Risk measurement.} Before designing a suite, it is helpful to decide what type of risk the suite will measure\cite{ibrahim2024beyond}:
        \begin{enumerate}
            \item Absolute risk: the direct risk from the AI system.
            \item Marginal risk: the increase in risk from the AI system, considering other technologies and variables (e.g., information widely available).
            \item Residual risk: the increase in risk from the system, assuming the safety safeguards put in place are not bypassed.
        \end{enumerate}
    Each of the above requires different evaluations and most likely would also affect other parameters in this list.
    \item \textit{Included evaluation types.}\footnote{For example, CTF-style evaluations are one type of evaluation, while MCQs are another.} Suites can range from being composed of a single evaluation type repeated to increase result confidence, to consisting of multiple evaluation types examining different capabilities, all relevant to a specific risk scenario.
    \item \textit{Quality and quantity.} Given limited resources, a trade-off may arise between evaluations' quality and quantity. Neither extreme provides sufficient information: a single high-quality evaluation rarely yields meaningful conclusions, while numerous low-quality evaluations are similarly uninformative in most cases. The decision-making process should help find an optimal balance between these extremes.
    \item \textit{Coverage and scope.} Suites can aim to be comprehensive regarding a critical capability, e.g., testing all possible categories of memory exploits. Alternatively, they may opt for limited evaluations across multiple areas of focus, emphasizing the assessment of various aspects within a specific risk scenario.
    \item \textit{Randomization and adaptability.} It is possible to design suites with a dynamically changing composition of evaluations, e.g., randomly or in response to the AI system's performance. This can increase the amount of information derived from the suite, but should be carefully considered as it might also cause the suite to be inconsistent or unreliable. It should also be noted that the additional value from variations or adaptations to the suite depends heavily on the extent to which the added/removed evaluations differ from one another.
    \item \textit{Expected shelf life.} As AI capabilities are increasing rapidly, many benchmarks are quickly being surpassed. It is important to estimate the suite's period of relevance and derive the necessary implications for the included evaluations. This can include the planned efforts to maintain and upkeep the suite, or a deadline for when a newly created suite should be ready.
\end{enumerate}

To emphasize, all of the above should be dictated by the encompassing decision making process and AI risk mapping that we described in Section~\ref{sec:threat_to_eval}. This directly affects the choices and adjustments to be made regarding the above parameters. Furthermore, there are many additional parameters beyond those described here. For example, the suite could either be public or private, and may include documentation. These parameters and others should be given careful thought.

\section{Conclusions}
Designing effective evaluations is a difficult task, and in the context of safety evaluations, it is also crucial for supporting high-stake decision-making processes. The combination of complex AI models and vague definitions of (possibly overlapping) capabilities poses additional obstacles.
In this draft safety framework paper, we first outlined the fundamental steps of AI risk mapping. These include defining the decision making framework, specifying the threat model and risk scenario, identifying the critical capabilities requiring evaluation, and establishing specific thresholds for these capabilities. 

We exemplify how the design of a specific evaluation, though seemingly straightforward, involves several non-trivial challenges. After showcasing common pitfalls, we suggest best practices and guidelines for constructing a useful evaluation. The list of desired properties for an evaluation includes a connection to a risk scenario, exclusion from the training set, clear difficulty measure, subject focus, high signal density, and coherent scoring method. Next, when these evaluations are aggregated to form an evaluation suite, we provide an overview of attributes and principles that should guide the development of most evaluation suites: coverage, difficulty level, scoring method and coverage overlap.

Furthermore, for both evaluations and evaluation suites, we list additional parameters that are worth considering. These parameters are more context-dependent and adjustable, and should be used according to specific needs. Some of these properties have specific trade-offs, where the optimum depends on the context. It is possible that for some suites of evaluations, overlooking these parameters could still result in an excellent evaluation suite.

To conclude, the art of designing evaluations and evaluation suites has many moving parts and considerations, and is more nuanced than it appears.
In this work, we have provided guidelines and tools to help practitioners improve their evaluation design process. This work is part of the first steps towards setting a community standard for safety evaluations. We believe it is an elusive problem, and further research is needed in this field.
We already follow these principles in our work on evaluations, and see value in sharing our approach.
Further research, such as exploring these best practices through empirical analysis or by specific case studies, is left for future work.

\bibliographystyle{plain}

\begin{thebibliography}{10}

\bibitem{abramovich2024enigma}
Talor Abramovich, Meet Udeshi, Minghao Shao, Kilian Lieret, Haoran Xi, Kimberly Milner, Sofija Jancheska, John Yang, Carlos~E Jimenez, Farshad Khorrami, et~al.
\newblock Enigma: Enhanced interactive generative model agent for ctf challenges.
\newblock {\em arXiv preprint arXiv:2409.16165}, 2024.

\bibitem{ahn2024large}
Janice Ahn, Rishu Verma, Renze Lou, Di~Liu, Rui Zhang, and Wenpeng Yin.
\newblock Large language models for mathematical reasoning: Progresses and challenges.
\newblock {\em arXiv preprint arXiv:2402.00157}, 2024.

\bibitem{anderljung2023frontier}
Markus Anderljung, Joslyn Barnhart, Anton Korinek, Jade Leung, Cullen O'Keefe, Jess Whittlestone, Shahar Avin, Miles Brundage, Justin Bullock, Duncan Cass-Beggs, et~al.
\newblock Frontier ai regulation: Managing emerging risks to public safety.
\newblock {\em arXiv preprint arXiv:2307.03718}, 2023.

\bibitem{Ant_RSP}
{Anthropic}.
\newblock Anthropic - responsible scaling policy.
\newblock \url{https://assets.anthropic.com/m/24a47b00f10301cd/original/Anthropic-Responsible-Scaling-Policy-2024-10-15.pdf/}, 2024.
\newblock Accessed: 2024-11-10.

\bibitem{Claude_3.5_Sonnet}
{Anthropic}.
\newblock Claude 3.5 sonnet.
\newblock \url{https://www.anthropic.com/news/claude-3-5-sonnet}, 2024.
\newblock Accessed: 2024-11-10.

\bibitem{We_need_a_Science_of_Evals}
{Appolo Research}.
\newblock We need a science of evals.
\newblock \url{https://www.apolloresearch.ai/blog/we-need-a-science-of-evals}, 2024.
\newblock Accessed: 2024-11-10.

\bibitem{baidoo2023education}
David Baidoo-Anu and Leticia~Owusu Ansah.
\newblock Education in the era of generative artificial intelligence (ai): Understanding the potential benefits of chatgpt in promoting teaching and learning.
\newblock {\em Journal of AI}, 7(1):52--62, 2023.

\bibitem{bengio2024managing}
Yoshua Bengio, Geoffrey Hinton, Andrew Yao, Dawn Song, Pieter Abbeel, Trevor Darrell, Yuval~Noah Harari, Ya-Qin Zhang, Lan Xue, Shai Shalev-Shwartz, et~al.
\newblock Managing extreme ai risks amid rapid progress.
\newblock {\em Science}, 384(6698):842--845, 2024.

\bibitem{brown2020language}
Tom~B Brown.
\newblock Language models are few-shot learners.
\newblock {\em arXiv preprint arXiv:2005.14165}, 2020.

\bibitem{burden2024evaluatingai}
John Burden.
\newblock Evaluating ai evaluation: Perils and prospects.
\newblock {\em arXiv preprint arXiv:2407.09221}, 2024.

\bibitem{picoCTF}
{Carnegie Mellon University}.
\newblock picoctf.
\newblock \url{https://picoctf.org/,}, 2024.
\newblock Accessed: 2024-11-10.

\bibitem{feffer2024red}
Michael Feffer, Anusha Sinha, Wesley~H Deng, Zachary~C Lipton, and Hoda Heidari.
\newblock Red-teaming for generative ai: Silver bullet or security theater?
\newblock In {\em Proceedings of the AAAI/ACM Conference on AI, Ethics, and Society}, volume~7, pages 421--437, 2024.

\bibitem{ganguli2023challenges}
Deep Ganguli, Nicholas Schiefer, Marina Favaro, and Jack Clark.
\newblock Challenges in evaluating {AI} systems, 2023.

\bibitem{GDM_FSF}
{Google Deep Mind}.
\newblock Introducing the frontier safety framework.
\newblock \url{https://deepmind.google/discover/blog/introducing-the-frontier-safety-framework/}, 2024.
\newblock Accessed: 2024-11-10.

\bibitem{hendrycks2023overview}
Dan Hendrycks, Mantas Mazeika, and Thomas Woodside.
\newblock An overview of catastrophic ai risks.
\newblock {\em arXiv preprint arXiv:2306.12001}, 2023.

\bibitem{Devising_ML_Metrics}
{Hendrycks, Dan and Woodside, Thomas}.
\newblock Devising ml metrics.
\newblock \url{https://www.safe.ai/blog/devising-ml-metrics}, 2024.
\newblock Accessed: 2024-11-10.

\bibitem{ibrahim2024beyond}
Lujain Ibrahim, Saffron Huang, Lama Ahmad, and Markus Anderljung.
\newblock Beyond static ai evaluations: advancing human interaction evaluations for llm harms and risks.
\newblock {\em arXiv preprint arXiv:2405.10632}, 2024.

\bibitem{ivanova2024bestresearchpractices}
Anna~A. Ivanova.
\newblock Toward best research practices in ai psychology.
\newblock {\em arXiv preprint arXiv:2312.01276}, 2024.

\bibitem{jason_wei_evals}
{Jason Wei}.
\newblock Successful language model evals.
\newblock \url{https://www.jasonwei.net/blog/evals}, 2024.
\newblock Accessed: 2024-11-10.

\bibitem{koessler2024risk}
Leonie Koessler, Jonas Schuett, and Markus Anderljung.
\newblock Risk thresholds for frontier ai.
\newblock {\em arXiv preprint arXiv:2406.14713}, 2024.

\bibitem{kolasani2023optimizing}
Saydulu Kolasani.
\newblock Optimizing natural language processing, large language models (llms) for efficient customer service, and hyper-personalization to enable sustainable growth and revenue.
\newblock {\em Transactions on Latest Trends in Artificial Intelligence}, 4(4), 2023.

\bibitem{kushner2013real}
David Kushner.
\newblock The real story of stuxnet.
\newblock {\em ieee Spectrum}, 50(3):48--53, 2013.

\bibitem{lee2023benefits}
Peter Lee, Sebastien Bubeck, and Joseph Petro.
\newblock Benefits, limits, and risks of gpt-4 as an ai chatbot for medicine.
\newblock {\em New England Journal of Medicine}, 388(13):1233--1239, 2023.

\bibitem{lehman2019surprisingcreativitydigitalevolution}
Joel Lehman, Jeff Clune, Dusan Misevic, Christoph Adami, Lee Altenberg, Julie Beaulieu, Peter~J. Bentley, Samuel Bernard, Guillaume Beslon, David~M. Bryson, Patryk Chrabaszcz, Nick Cheney, Antoine Cully, Stephane Doncieux, Fred~C. Dyer, Kai~Olav Ellefsen, Robert Feldt, Stephan Fischer, Stephanie Forrest, Antoine Frénoy, Christian Gagné, Leni~Le Goff, Laura~M. Grabowski, Babak Hodjat, Frank Hutter, Laurent Keller, Carole Knibbe, Peter Krcah, Richard~E. Lenski, Hod Lipson, Robert MacCurdy, Carlos Maestre, Risto Miikkulainen, Sara Mitri, David~E. Moriarty, Jean-Baptiste Mouret, Anh Nguyen, Charles Ofria, Marc Parizeau, David Parsons, Robert~T. Pennock, William~F. Punch, Thomas~S. Ray, Marc Schoenauer, Eric Shulte, Karl Sims, Kenneth~O. Stanley, François Taddei, Danesh Tarapore, Simon Thibault, Westley Weimer, Richard Watson, and Jason Yosinski.
\newblock The surprising creativity of digital evolution: A collection of anecdotes from the evolutionary computation and artificial life research communities.
\newblock {\em arXiv preprint arXiv:1803.03453}, 2019.

\bibitem{li2024wmdp}
Nathaniel Li, Alexander Pan, Anjali Gopal, Summer Yue, Daniel Berrios, Alice Gatti, Justin~D Li, Ann-Kathrin Dombrowski, Shashwat Goel, Long Phan, et~al.
\newblock The wmdp benchmark: Measuring and reducing malicious use with unlearning.
\newblock {\em arXiv preprint arXiv:2403.03218}, 2024.

\bibitem{liang2022holistic}
Percy Liang, Rishi Bommasani, Tony Lee, Dimitris Tsipras, Dilara Soylu, Michihiro Yasunaga, Yian Zhang, Deepak Narayanan, Yuhuai Wu, Ananya Kumar, et~al.
\newblock Holistic evaluation of language models.
\newblock {\em arXiv preprint arXiv:2211.09110}, 2022.

\bibitem{Responsible_Scaling_Policies}
{METR}.
\newblock Responsible scaling policies (rsps).
\newblock \url{https://metr.org/blog/2023-09-26-rsp/}, 2023.
\newblock Accessed: 2024-11-10.

\bibitem{OpenAI_Preparedness}
{OpenAI}.
\newblock Openai - preparedness framework (beta).
\newblock \url{https://cdn.openai.com/openai-preparedness-framework-beta.pdf}, 2023.
\newblock Accessed: 2024-11-10.

\bibitem{Dall-E-3}
{OpenAI}.
\newblock Dall·e 3.
\newblock \url{https://openai.com/index/dall-e-3/}, 2024.
\newblock Accessed: 2024-11-10.

\bibitem{Hello_GPT-4o}
{OpenAI}.
\newblock Hello gpt-4o.
\newblock \url{https://openai.com/index/hello-gpt-4o/}, 2024.
\newblock Accessed: 2024-11-10.

\bibitem{openai2019rubiks}
OpenAI, Ilge Akkaya, Marcin Andrychowicz, Maciek Chociej, Mateusz Litwin, Bob McGrew, Arthur Petron, Alex Paino, Matthias Plappert, Raphael Powell, Glenn abd~Ribas, Jonas Schneider, Nikolas Tezak, Jerry Tworek, Peter Welinder, Lilian Weng, Qiming Yuan, Wojciech Zaremba, and Lei Zhang.
\newblock Solving rubik's cube with a robot hand.
\newblock {\em arXiv preprint arXiv:1910.07113}, 2019.

\bibitem{CSAW_CTF_2019_Quals}
{Osiris}.
\newblock Csaw ctf 2019 quals.
\newblock \url{https://github.com/osirislab/CSAW-CTF-2019-Quals}, 2019.
\newblock Accessed: 2024-11-10.

\bibitem{CSAW_CTF_2022_Finals}
{Osiris}.
\newblock Csaw ctf 2022 finals.
\newblock \url{https://github.com/osirislab/CSAW-CTF-2022-Finals}, 2022.
\newblock Accessed: 2024-11-10.

\bibitem{ezMaze}
{Osiris}.
\newblock ezmaze.
\newblock \url{https://github.com/osirislab/CSAW-CTF-2022-Quals/tree/master/misc/ezMaze}, 2022.
\newblock Accessed: 2024-11-10.

\bibitem{CSAW_CTF_2023_Finals}
{Osiris}.
\newblock Csaw ctf 2023 finals.
\newblock \url{https://github.com/osirislab/CSAW-CTF-2023-Finals}, 2023.
\newblock Accessed: 2024-11-10.

\bibitem{urkel}
{Osiris}.
\newblock urkel.
\newblock \url{https://github.com/osirislab/CSAW-CTF-2023-Finals/tree/main/misc/urkel}, 2023.
\newblock Accessed: 2024-11-10.

\bibitem{phuong2024evaluating}
Mary Phuong, Matthew Aitchison, Elliot Catt, Sarah Cogan, Alexandre Kaskasoli, Victoria Krakovna, David Lindner, Matthew Rahtz, Yannis Assael, Sarah Hodkinson, et~al.
\newblock Evaluating frontier models for dangerous capabilities.
\newblock {\em arXiv preprint arXiv:2403.13793}, 2024.

\bibitem{sainz2023nlp}
Oscar Sainz, Jon~Ander Campos, Iker Garc{\'\i}a-Ferrero, Julen Etxaniz, Oier~Lopez de~Lacalle, and Eneko Agirre.
\newblock Nlp evaluation in trouble: On the need to measure llm data contamination for each benchmark.
\newblock {\em arXiv preprint arXiv:2310.18018}, 2023.

\bibitem{shao2024empirical}
Minghao Shao, Boyuan Chen, Sofija Jancheska, Brendan Dolan-Gavitt, Siddharth Garg, Ramesh Karri, and Muhammad Shafique.
\newblock An empirical evaluation of llms for solving offensive security challenges.
\newblock {\em arXiv preprint arXiv:2402.11814}, 2024.

\bibitem{shao2024nyudataset}
Minghao Shao, Sofija Jancheska, Meet Udeshi, Brendan Dolan-Gavitt, Haoran Xi, Kimberly Milner, Boyuan Chen, Max Yin, Siddharth Garg, Prashanth Krishnamurthy, et~al.
\newblock Nyu ctf dataset: A scalable open-source benchmark dataset for evaluating llms in offensive security.
\newblock {\em arXiv preprint arXiv:2406.05590}, 2024.

\bibitem{shevlane2023model}
Toby Shevlane, Sebastian Farquhar, Ben Garfinkel, Mary Phuong, Jess Whittlestone, Jade Leung, Daniel Kokotajlo, Nahema Marchal, Markus Anderljung, Noam Kolt, et~al.
\newblock Model evaluation for extreme risks.
\newblock {\em arXiv preprint arXiv:2305.15324}, 2023.

\bibitem{slattery2024ai}
Peter Slattery, Alexander~K Saeri, Emily~AC Grundy, Jess Graham, Michael Noetel, Risto Uuk, James Dao, Soroush Pour, Stephen Casper, and Neil Thompson.
\newblock The ai risk repository: A comprehensive meta-review, database, and taxonomy of risks from artificial intelligence.
\newblock {\em arXiv preprint arXiv:2408.12622}, 2024.

\bibitem{tang2024any}
Zineng Tang, Ziyi Yang, Chenguang Zhu, Michael Zeng, and Mohit Bansal.
\newblock Any-to-any generation via composable diffusion.
\newblock {\em Advances in Neural Information Processing Systems}, 36, 2024.

\bibitem{Advanced_AI_evaluations_at_AISI:_May_update}
{UK AISI}.
\newblock Advanced ai evaluations at aisi: May update.
\newblock \url{https://www.aisi.gov.uk/work/advanced-ai-evaluations-may-update}, 2024.
\newblock Accessed: 2024-11-10.

\bibitem{wan2024cyberseceval}
Shengye Wan, Cyrus Nikolaidis, Daniel Song, David Molnar, James Crnkovich, Jayson Grace, Manish Bhatt, Sahana Chennabasappa, Spencer Whitman, Stephanie Ding, et~al.
\newblock Cyberseceval 3: Advancing the evaluation of cybersecurity risks and capabilities in large language models.
\newblock {\em arXiv preprint arXiv:2408.01605}, 2024.

\bibitem{wan2023kelly}
Yixin Wan, George Pu, Jiao Sun, Aparna Garimella, Kai-Wei Chang, and Nanyun Peng.
\newblock " kelly is a warm person, joseph is a role model": Gender biases in llm-generated reference letters.
\newblock {\em arXiv preprint arXiv:2310.09219}, 2023.

\bibitem{weidinger2024holisticsafetyresponsibilityevaluations}
Laura Weidinger, Joslyn Barnhart, Jenny Brennan, Christina Butterfield, Susie Young, Will Hawkins, Lisa~Anne Hendricks, Ramona Comanescu, Oscar Chang, Mikel Rodriguez, Jennifer Beroshi, Dawn Bloxwich, Lev Proleev, Jilin Chen, Sebastian Farquhar, Lewis Ho, Iason Gabriel, Allan Dafoe, and William Isaac.
\newblock Holistic safety and responsibility evaluations of advanced ai models, 2024.

\bibitem{xu2024vasa}
Sicheng Xu, Guojun Chen, Yu-Xiao Guo, Jiaolong Yang, Chong Li, Zhenyu Zang, Yizhong Zhang, Xin Tong, and Baining Guo.
\newblock Vasa-1: Lifelike audio-driven talking faces generated in real time.
\newblock {\em arXiv preprint arXiv:2404.10667}, 2024.

\bibitem{yan2024protecting}
Biwei Yan, Kun Li, Minghui Xu, Yueyan Dong, Yue Zhang, Zhaochun Ren, and Xiuzhen Cheng.
\newblock On protecting the data privacy of large language models (llms): A survey.
\newblock {\em arXiv preprint arXiv:2403.05156}, 2024.

\bibitem{zhang2024cybench}
Andy~K Zhang, Neil Perry, Riya Dulepet, Eliot Jones, Justin~W Lin, Joey Ji, Celeste Menders, Gashon Hussein, Samantha Liu, Donovan Jasper, et~al.
\newblock Cybench: A framework for evaluating cybersecurity capabilities and risk of language models.
\newblock {\em arXiv preprint arXiv:2408.08926}, 2024.

\bibitem{zhang2024llmeval}
Yue Zhang, Ming Zhang, Haipeng Yuan, Shichun Liu, Yongyao Shi, Tao Gui, Qi~Zhang, and Xuanjing Huang.
\newblock Llmeval: A preliminary study on how to evaluate large language models.
\newblock In {\em Proceedings of the AAAI Conference on Artificial Intelligence}, volume~38, pages 19615--19622, 2024.

\end{thebibliography}

\end{document}